\documentclass[conference]{IEEEtran}
\IEEEoverridecommandlockouts
\usepackage{cite}
\usepackage{float}
\usepackage{hyperref}
\usepackage{url}
\usepackage{amsmath,amssymb,amsfonts}
\usepackage{algorithmic}
\usepackage{graphicx}
\usepackage{textcomp}
\usepackage{xcolor}
\def\BibTeX{{\rm B\kern-.05em{\sc i\kern-.025em b}\kern-.08em
    T\kern-.1667em\lower.7ex\hbox{E}\kern-.125emX}}
\begin{document}

\title{Self-Supervised Real-Time Tracking of Military Vehicles in Low-FPS UAV Footage\\
}

\author{
    \IEEEauthorblockN{
        Markiyan Kostiv\textsuperscript{1,2},
        Anatolii Adamovskyi\textsuperscript{3},
        Yevhen Cherniavskyi\textsuperscript{1},\\
        Mykyta Varenyk\textsuperscript{2},
        Ostap Viniavskyi\textsuperscript{2},
        Igor Krashenyi\textsuperscript{2},
        Oles Dobosevych\textsuperscript{2}
    }
    \IEEEauthorblockA{
        \textsuperscript{1}\textit{CIDTD, MoD Ukraine}, Kyiv, Ukraine \\
        \textsuperscript{2}\textit{Applied Sciences Faculty, Ukrainian Catholic University}, Lviv, Ukraine \\
        \textsuperscript{3} Independent Resarcher, Kyiv, Ukraine \\
        Emails: \{m.kostiv, varenyk.pn, viniavskyi, igor.krashenyi, dobosevych\}@ucu.edu.ua,\\
        adamovskyi.anatolii@gmail.com
    }
}

\maketitle

\begin{abstract}
Multi-object tracking (MOT) aims to maintain consistent identities of objects across video frames. Associating objects in low-frame-rate videos captured by moving unmanned aerial vehicles (UAVs) in actual combat scenarios is complex due to rapid changes in object appearance and position within the frame. The task becomes even more challenging due to image degradation caused by cloud video streaming and compression algorithms. We present how instance association learning from single-frame annotations can overcome these challenges. We show that global features of the scene provide crucial context for low-fps instance association, allowing our solution to be robust to distractors and gaps in detections. We also demonstrate that such a tracking approach maintains high association quality even when reducing the input image resolution and latent representation size for faster inference. Finally, we present a benchmark dataset of annotated military vehicles collected from publicly available data sources. This paper was initially presented at the NATO Science and Technology Organization Symposium
(ICMCIS) organized by the Information Systems Technology (IST)Scientific and Technical
Committee, IST-209-RSY – the ICMCIS, held in Oeiras, Portugal, 13-14 May 2025.
\end{abstract}


\section{Introduction}
Significant cost reductions and increased effectiveness of UAVs led to the saturation of the battlefield with drones capable of producing high-quality video surveillance data at a scale never seen before. This shift enabled constant monitoring of a frontline through video streams. Multiple reconnaissance UAVs, deployed on a rotational basis, provide full battlefield coverage to a tactical depth and beyond. This greatly increases situational awareness and is crucial for the initial steps of an effective kill chain.

The abundance of real-time reconnaissance signals poses an additional challenge: swift and effective conversion of raw intelligence data into actionable intelligence items. Sole reliance on human operators responsible for monitoring multiple video streams and identifying military assets can lead to decreased efficiency and missed relevant activity of an adversary — an undesirable situation in a combat scenario.

An automation of military assets identification from real-time UAV video streams is an effective and scalable solution to the challenges introduced by the increased sheer volume of video surveillance data available. Object detection-based approaches alleviate the increased burden of extracting relevant signals from video streams from UAVs, acting as a copilot for an intelligence officer responsible for data processing and monitoring. However, sole object detection on a single frame from a video oftentimes doesn't produce actionable intelligence items. For efficient battlefield monitoring, operating with unique military assets is important - a task that object detection alone can't solve. Multiple object tracking (MOT), in addition to object detection, is an effective solution to the task of military assets identification from video.

Performing surveillance and target acquisition with neural networks in real-time video streams—particularly in the cloud—requires significant computational resources. Although raw video sources often capture footage at high frame rates (30 frames per second (FPS) or more), analyzing every frame is both -- resource-intensive and latency-prone. As a result, these systems typically decode and process only a subset of frames. Under these constraints, low-fps (around 5 FPS) MOT is essential for distinguishing unique targets and reliably filtering out distractors (false positives).

\subsection{Low-fps multiple object tracking}

Low-fps MOT is a challenging problem in computer vision due to the rapid changes in object appearance and the difficulty of accurately predicting future object positions using motion models \cite{kalman1960, 1eurofilter}. Although some high-fps methods assume relatively small displacements of objects between consecutive frames, low-fps videos introduce significant temporal gaps that make such assumptions irrelevant. This challenge is compounded by a moving camera's additional complexities, particularly when footage is captured from a UAV \cite{uav_tracking}. The camera can undergo substantial viewpoint shifts in these scenarios, making both simple motion models and overlap-based data-association strategies ineffective.

In high-fps tracking approaches, Intersection-over-Union (IoU) matching, often paired with a simple motion model, provides a reliable initial association step \cite{sort, deepsort, bytetrack}. Methods like ByteTrack \cite{bytetrack}, or BoostTrack\cite{boostTrack} leverage the overlap of bounding boxes in consecutive frames and basic kinematic predictions to link detections. These techniques rely on the assumption that objects move only slightly between frames, enabling both IoU-based comparison and short-term velocity estimation. Consequently, bounding boxes of the same object in consecutive frames are expected to overlap considerably, while simple motion extrapolation can further reduce the search space for potential matches.

However, when the time gap between processed frames increases and the camera’s viewpoint changes rapidly, the effectiveness of IoU-based and motion-model-based approaches is significantly reduced \cite{occlusion_and_motion}. Bounding boxes may show little to no overlap even for the same target, and any motion estimates derived from high-fps assumptions can deviate considerably from the actual trajectory. This leads to problems such as frequent identity switches, missed associations, or redundant new track initializations \cite{uav_tracking}. These problems persist in footage captured by UAVs, where the apparent movement of an object is influenced by both its motion and the rapid movements of the camera.

Recent multi-object tracking approaches \cite{deepsort, meinhardt2022trackformermultiobjecttrackingtransformers} demonstrate a strong ability to learn associations based on the latent representation of objects, performing well even in low-fps scenarios. However, these methods rely on precise annotations with tracks, which is time-consuming and resource-intensive, particularly for large-scale datasets. To address this challenge, we propose leveraging single-frame annotations instead, as they provide a shorter and more efficient path to dataset diversity and improved detection performance. Single-frame datasets capture various object appearances, viewpoints, and contexts without the overhead of temporal annotation. Inspired by MASA self-supervised training \cite{li2024matchingsegmenting}, we demonstrate that instance association can be learned in a self-supervised manner directly from these single-frame datasets. This approach simplifies the data collection process and enables robust tracking capabilities by leveraging the inherent diversity of single-frame annotations to improve association learning.

Furthermore, we demonstrate that in low-fps UAV footage, the spatial locations of objects within the scene and the feature levels extracted from them are crucial for successful tracking. Rapid viewpoint shifts and varying altitudes of UAV cameras introduce significant spatial context changes, making it critical to learn representations that incorporate both local object features and global scene-level context. These challenges are further exacerbated by the temporal gaps in low-fps scenarios, where effective spatial- and feature-level modeling becomes essential to maintain consistent associations across frames.

Finally, we present a benchmark dataset of annotated military vehicles for MOT collected from publicly available data sources. The data is captured from various UAVs, encompassing diverse viewpoints, altitudes, and operational scenarios. This dataset offers a strong basis for advancing research in low-fps multi-object tracking and lays the groundwork for developing robust tracking solutions in challenging real-world combat scenarios.
\newpage

\section{Related works}

Frame-to-frame annotations for multi-object tracking are inherently challenging due to the necessity of precise re-identification (ReID) across consecutive frames amidst frequent occlusions, varying object appearances, and diverse viewing angles. Additionally, manual labeling processes introduce natural "bounding box jitter," where slight discrepancies in annotated bounding boxes can negatively impact evaluation metrics by reducing precision and accuracy. Thus, our work leverages data annotated for single-frame detection. This constraint rules out approaches that heavily rely on dense inter-frame labels and motivates methods that can cope with sparse or missing temporal annotations. In addition, low-FPS video introduces large inter-frame displacements, which can break standard assumptions about smooth or linear motion.

\subsection{Tracking-by-Detection with Feature Stabilization}\label{AA}
One of the most popular paradigms for MOT is Tracking by detection.  In this approach, a detector processes each frame independently to produce bounding boxes for candidate objects. A separate module then links these detections across frames to form consistent object trajectories. This paradigm belongs to a family of Re-Identification (Re-ID) trackers that leverage both appearance embeddings and predictive motion modeling to address the challenges inherent in object tracking, such as occlusion, motion blur, and dynamic environments.

DeepSORT\cite{deepsort} employs a Kalman filter (KF)\cite{kalman1960} to predict object bounding boxes in subsequent frames, assuming near-linear motion patterns. This predictive capability significantly reduces the search space for associating detections across frames. For Re-ID, DeepSORT utilizes a deep appearance descriptor to extract discriminative embeddings for each detected bounding box using a lightweight convolutional neural network (CNN). IoU is employed for initial geometric consistency during matching. By unifying predictive motion models and latent object representations, DeepSORT demonstrates robust performance in MOT tasks, even under challenging conditions such as occlusion and erratic object motion.

OC-SORT\cite{ocsort} is a more recent representative of the SORT family that tries to improve motion predictions done by the KF when trajectories are non-linear or when displacements between frames are too large. Specifically, instead of solely relying on a bounding box from the KF, OC-SORT takes the latest observation after being untracked (“reactivation”), and the parameters of KF are updated.

FairMOT \cite{fairmot} tries to formulate the training process into a multi-task learning of object detection and ReID in a single network. This integration reduces inconsistencies between detection and feature embedding, a common challenge in tracking-by-detection methods, and makes the training process more efficient. FairMOT uses a center-based anchor-free detection mechanism CenterTrack \cite{centertrack} since the authors have identified that the use of multiple anchors near the same objects leads to performance degradation.

ByteTrack\cite{bytetrack} became well-known for achieving leading results in popular MOT benchmarks by maximizing detection recall while minimizing false positives. ByteTrack operates as a two-stage algorithm: 
\begin{itemize}
    \item High-Confidence Matching: filter detections by a high confidence threshold and perform standard IoU matching with existing tracks. This step ensures robust associations when the detector is confident.
    \item Low-Confidence Re-Association: Take unassigned detections above a lower threshold and try to match them with “lost” or unmatched tracks if the IoU is reasonable.
\end{itemize}

All of these methods tend to perform exceptionally well on high-frame-rate benchmarks. Still, they commonly rely on IoU in object ReID, which in low fps regimes leads to bounding boxes showing little to no overlap, even for the same target. Thus, these methods are prone to significant performance degradation in low fps conditions \cite{lowfps_assessment}. Additionally, most of them also rely on consecutive frame-to-frame annotations for training. 

\subsection{Frame-Rate-Insensitive methods}

Several algorithms aim to be robust in frame-rate-insensitive environments, among them ColTrack\cite{collab_tracking_fri}, APPTracker\cite{apptracker}, and FraMOT\cite{framot}. 

ColTrack is designed to handle sparse and/or irregular frame rates by maintaining a shared embedding across multiple frames. This strategy helps close large temporal gaps by comparing new detections against stored embeddings from a pool of historical frames. Historical information is iteratively refined using an Information Refinement Module (IRM). The IRM combines a removal branch to filter out irrelevant data and an addition branch to incorporate valuable temporal clues. Furthermore, a Tracking Object Consistency Loss (TOCLoss) which requires each tracking query to collect discriminative features from
other historical queries for the correct tracking. 

FraMOT aims to learn a unified model to track objects in videos with agnostic frame rates by introducing the Frame Rate Agnostic Association Module (FRAAM) and Periodic Training Scheme (PTS). FRAAM tries to address small IoU overlaps and train the appearance model to handle large displacements. Instead of assuming a fixed “one-frame-ahead” motion, parameter $\Delta t$ to account for the number of frames skipped. Adaptive thresholding is used to adjust bounding boxes. PTS purposefully varies the temporal spacing between frames in the training dataset to mimic different frame rates.

APPTracker tries to improve CenterTrack by introducing the Appearance-and-Prediction-Probability(APP) head. CenterTrack is implemented using optical flow, which attempts to predict motion using the last two consecutive frames and struggles when an object is absent in the previous frame. The APP head detects new objects but does not calculate displacement prediction. This has led to decreased identity switches and higher IDF1 scores in low-fps scenarios compared to CenterTrack.

Still, although some improvements were achieved in low-fps scenarios, these methods required complete MOT labeling. Besides, the methods followed the tracking-by-detection paradigm, which relies on motion models that try to predict an object's next location based on short-term previous trajectory. Meanwhile, an alternative method could be to have end-to-end learning of both detection and tracking simultaneously, which allows for shared feature representation for both tasks. 

\subsection{Transformer based methods}

After DETR\cite{detr} has introduced object queries and global attention, Trackformer\cite{meinhardt2022trackformermultiobjecttrackingtransformers} added track queries to reference a specific tracked instance in a previous frame. The encoder processes the image features, while the decoder attends to both object features and track queries. Global attention allows the Trackformer to scan the entire image when matching objects across frames. As a result, the model can more easily find and re-identify an object that has moved significantly between frames. 

MOTR\cite{motr} introduced persistent track queries that remain active across the entire video, storing each object’s features in a stronger memory module. This module retains track queries even if an object is temporarily occluded or not detected, significantly reducing ID switches because the same queries carry forward once the object reappears. However, if many frames pass or the sequence is especially long, managing all these active queries can become computationally demanding. MOTRv3 \cite{motrv3} addresses this with hierarchical memory and enhanced query management, where track queries are selectively stored (e.g., at “keyframes”), rather than for every single frame. This hierarchical approach allows the model to efficiently revisit past track states while avoiding the overhead of storing a full memory of every frame. MOTRv3 also refines how queries transition between “active” (definitely tracking an object) and “candidate” or “inactive” states.

One of the most recent developments is MeMOTR \cite{memotr}, which incorporates a long-term memory module that maintains a consistent representation. Memory updates use an exponential moving average with a tunable rate, ensuring gradual adaptation to temporal changes while retaining historical information. MeMOTR introduces Temporal Interaction Module (TIM), where adaptive aggregation mechanism fuses information from adjacent frames while prioritizing reliable features. In low-FPS scenarios, this aggregation can compensate for sparse temporal information by emphasizing stable features across time. 

\subsection{Self-Supervised Approaches}

Self-supervised approaches have gained traction in addressing the issue of costly inter-frame annotation. Most recent advancements include MASA \cite{li2024matchingsegmenting}, PCL \cite{self_supervised_mot_path_consistency}.

MASA bypasses the need for dense inter-frame labeling by leveraging quasi-dense similarity learning \cite{quasidense}, which establishes fine-grained correspondences between object regions across frames. Rather than relying on strict one-to-one bounding box matches, MASA samples region-level features in a quasi-dense manner, computing similarity scores that can handle significant appearance changes and detection gaps. These region-level matches help maintain track identities when frames are far apart, making MASA particularly well-suited for low-FPS scenarios.

Path Consistency Loss (PCL) [9] imposes consistency constraints over multiple “skipped” frames, training the model to link objects reliably across both short and long gaps. By learning to re-identify objects over varied temporal distances, PCL naturally adapts to low-FPS data, where smooth, frame-by-frame transitions are not guaranteed.

\section{Validation benchmark}

\subsection{Military Vehicle Classification}

We have developed a comprehensive validation benchmark specifically designed for low-FPS military MOT scenarios. This benchmark focuses on evaluating the performance of tracking algorithms under challenging conditions, such as low frame rates, occlusions, and varying object dynamics. The benchmark includes three distinct object classes, each representing critical elements commonly encountered in military environments. These classes are as follows:
\begin{itemize}
    \item \textbf{Heavily Armored} - entities that include air defense systems, artillery, tanks, infantry fighting vehicles (IFV), and multiple launch rocket systems (MLRS).
    \item \textbf{Lightly Armored or Unarmored} - militarized cars, infantry mobile vehicles, reconnaissance, and patrol vehicles.
    \item \textbf{Trucks} - fuel trucks, KUNG military trucks, general-purpose military trucks, etc.
\end{itemize}

\subsection{Benchmark Composition}
Data sources for the raw data collection were various Telegram channels, x.com, and YouTube channels. During the search, we focused on videos recorded using quadcopters or wing-type drones, providing views from high altitudes to cover large scenes. Our primary goal was to identify videos that depict two or more vehicles with changing angles, resolutions, and object positions that closely resemble real combat scenarios. 

The dataset contains:
\begin{itemize}
    \item \textbf{Videos:} 119 videos collected from diverse sources, with resolutions ranging from 480p to 4K, and analyzed durations between 7 seconds and 1:30 minutes. The videos are in mp4 and mov formats.
    \item \textbf{Frames:} 18421 
    \item \textbf{Number of bounding boxes:} 58897
    \item \textbf{Weather Diversity:} 14\% of the videos were recorded in winter, while 86\% were captured in summer 
\end{itemize}

\begin{figure}[H]
    \centering
    \includegraphics[width=\linewidth]{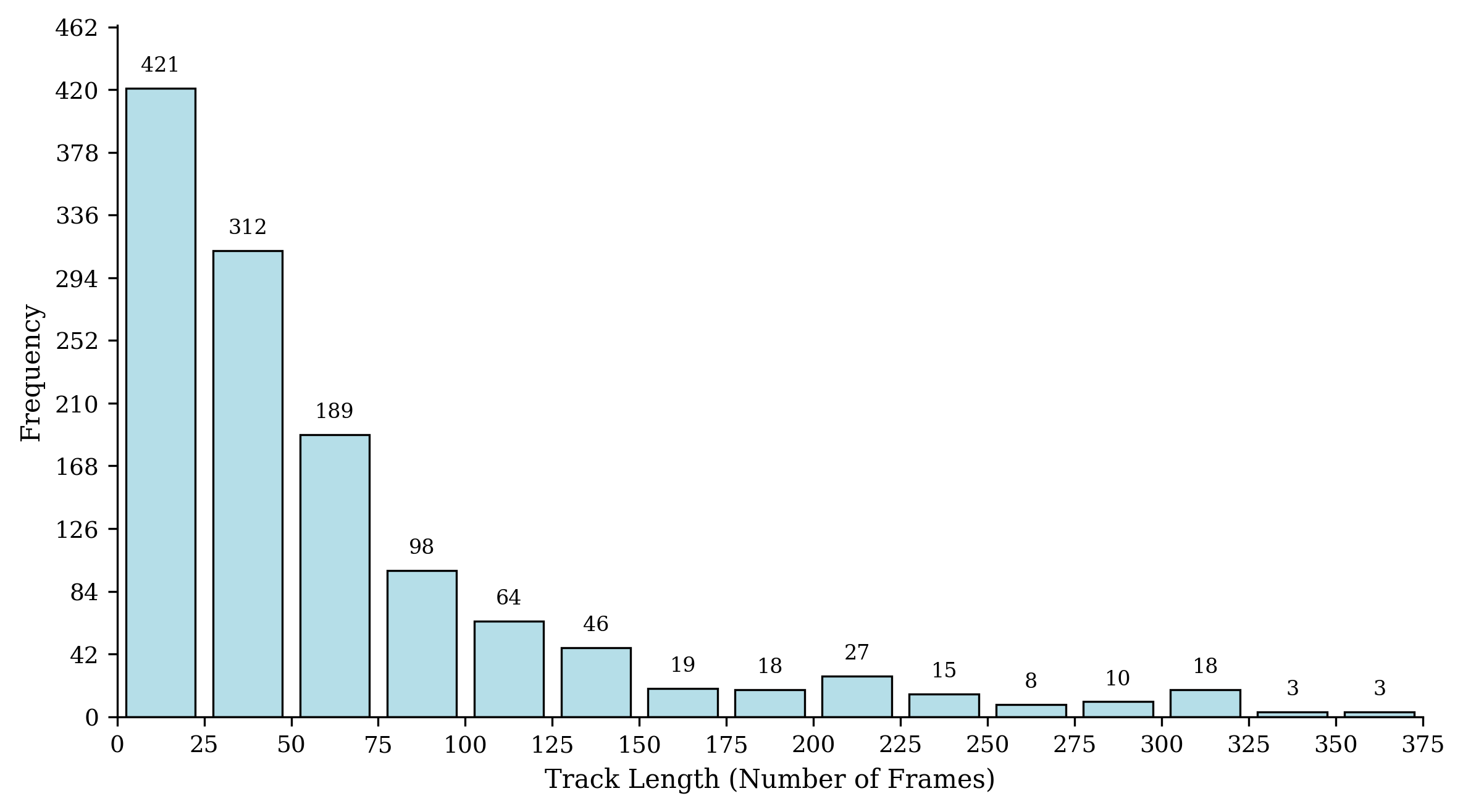}
    \caption{Track Length Distribution}
    \label{fig:enter-label}
\end{figure}

\begin{figure}[H]
    \centering
    \includegraphics[width=\linewidth]{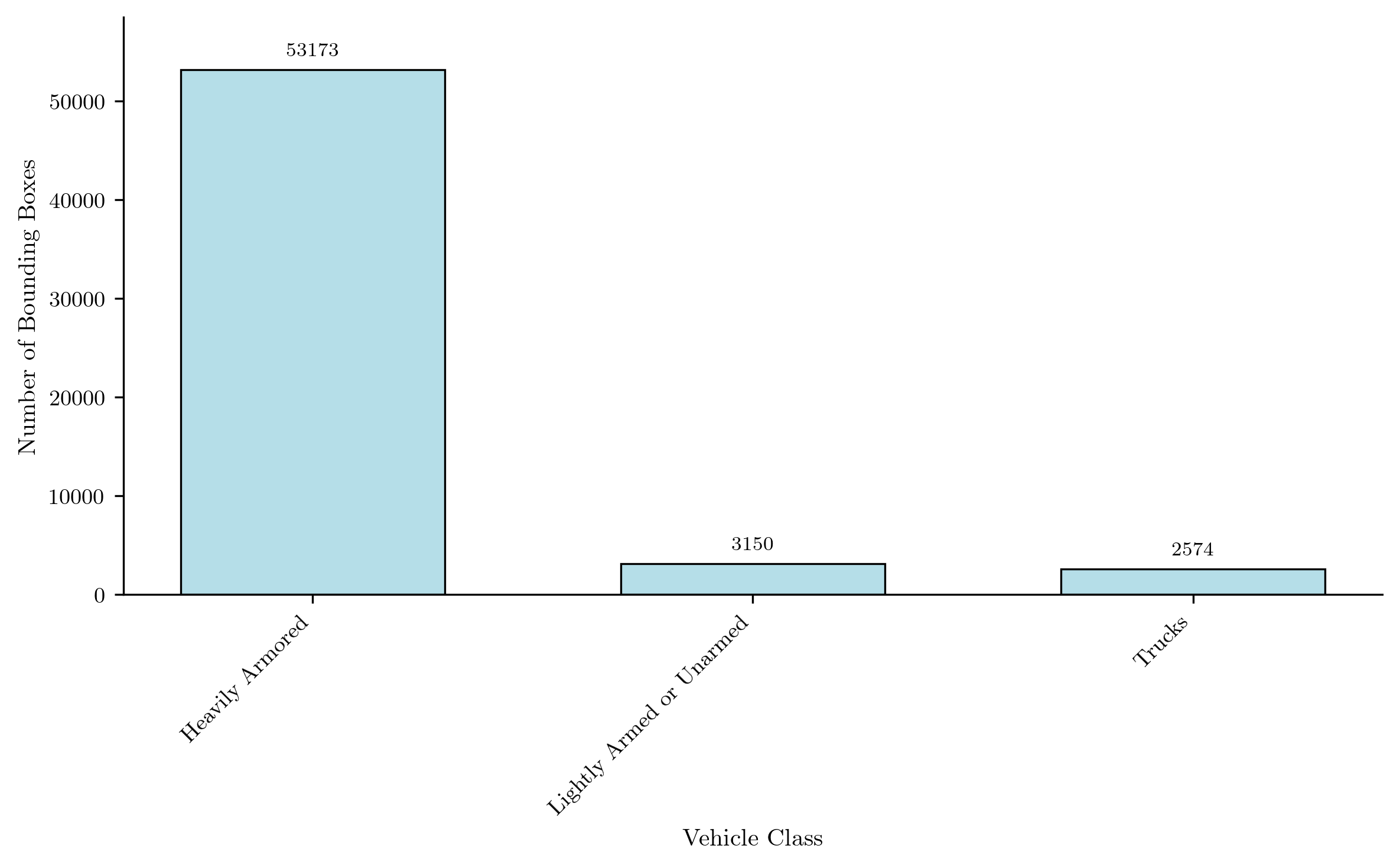}
    \caption{Classes distribution}
    \label{fig:enter-label}
\end{figure}

\begin{figure}[h!]
    \begin{tabular}{ccc}
        \includegraphics[width=0.5\linewidth] {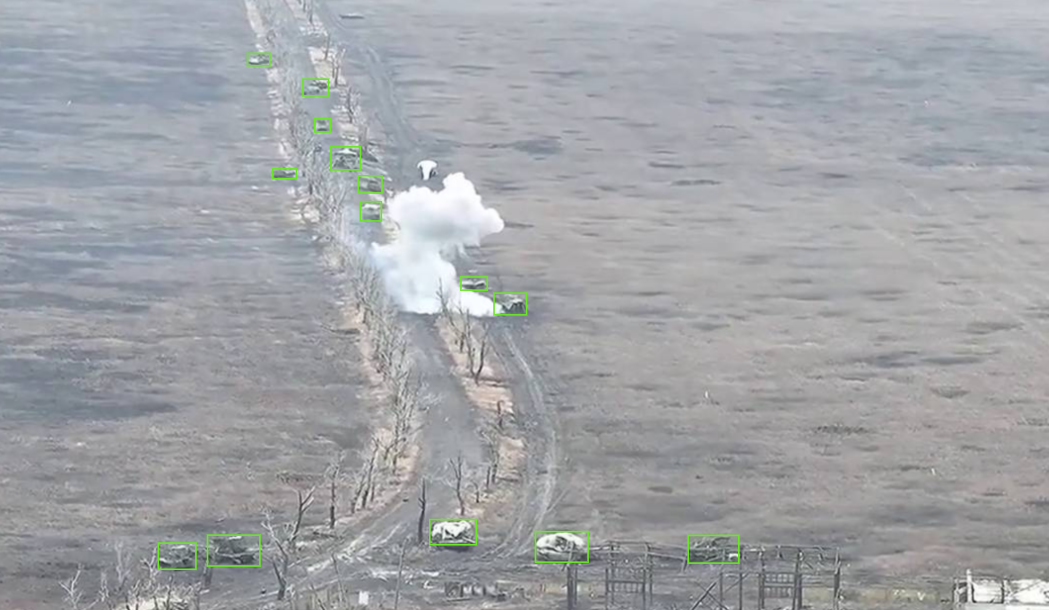} & 
        \includegraphics[width=0.52\linewidth]{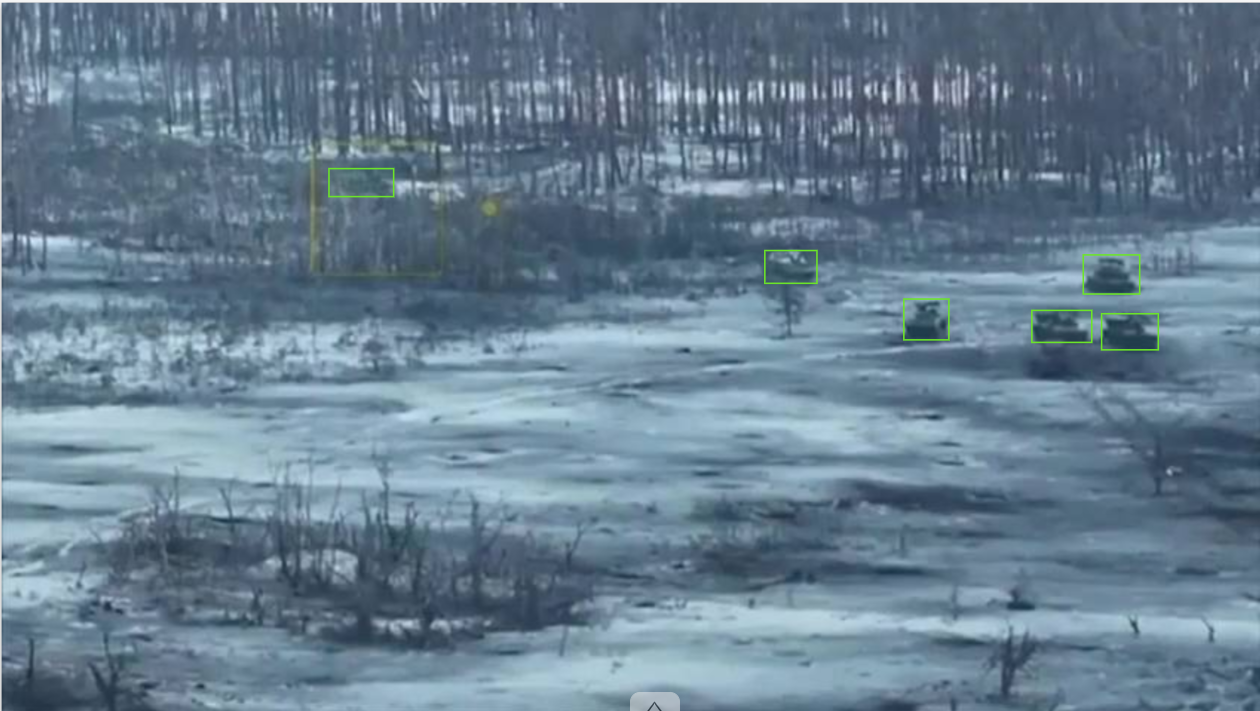} \\
        \includegraphics[width=0.5\linewidth]{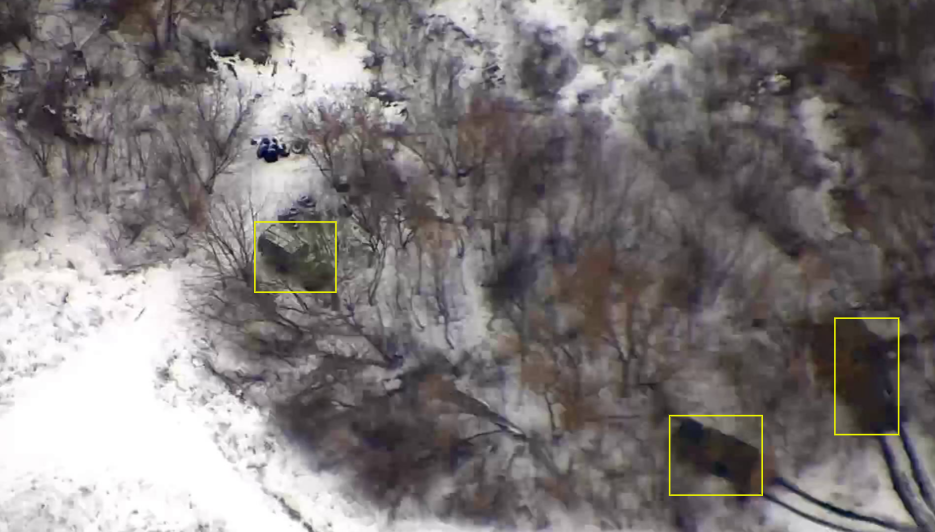} &
        \includegraphics[width=0.51\linewidth ]{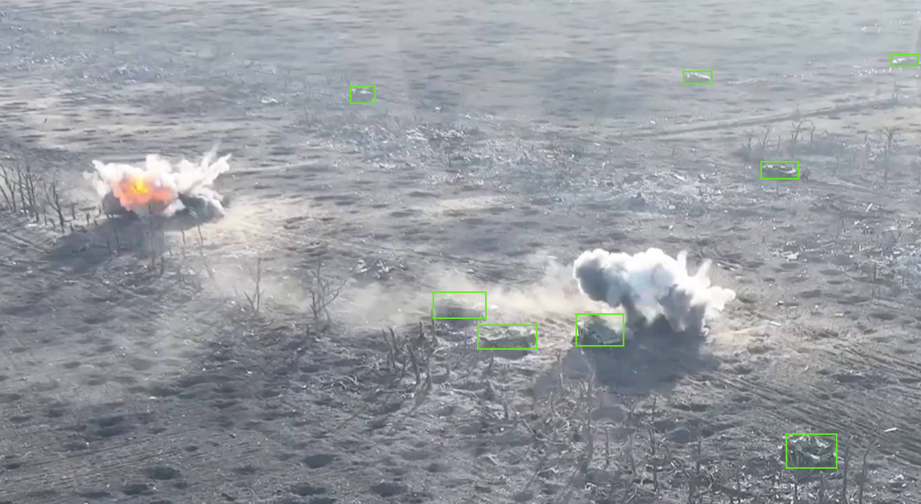} 
    
    \end{tabular}
     \caption{Examples of annotated videos}
\end{figure}

\subsection{Annotation Process}
The annotation team consisted of human experts from Ukrainian Catholic University who had prior exposure to Computer Vision. 
 
\begin{itemize}
    \item Tracks were annotated by identifying and labeling individual vehicles across frames and assigning unique IDs to maintain track consistency over time.
    \item Annotations included bounding boxes, object classes, and temporal links for vehicle tracking.
    \item Quality control measures were implemented, involving multiple reviews of annotated tracks by different team members to minimize labeling errors.
\end{itemize}

\section{Method}
We adopt the MASA framework \cite{li2024matchingsegmenting} for self-supervised, instance-level latent representation learning and matching from a single-frame dataset. We train the MASA model with a ResNet50 \cite{resnet} backbone by first pre-training it on a subset of the diverse, general-purpose SA-1B \cite{kirillov2023segment} dataset using pseudo-labels generated by Segment Anything Model (SAM) \cite{kirillov2023segment}, and then fine-tuning it on a smaller UAV footage dataset containing single-frame bounding box annotations of various military vehicle types. All experiments and results are based on the detections obtained from a binary object detection model trained on the same UAV footage dataset later used from MASA training. Inspired by MASA, we employ self-supervised contrastive learning technique \cite{khosla2021supervisedcontrastivelearning, li2022trackingthingwild} and an object prior distillation branch as an auxiliary task during training \cite{ren2016fasterrcnnrealtimeobject}. In the fine-tuning stage, we also retain frames with a single annotated object to reinforce positive pair learning by increasing the amount of single-object positive pairs and better aligning with real-world UAV scenarios where single-object tracking is common.

\section{Experiments and Discussion}

\subsection{Data Annotation Strategy for Training}

Accurate data annotation is essential for training tracking models, particularly in UAV-based military asset identification. We experimented with two annotation strategies: pseudo-labeling using the SAM, as suggested in the original MASA framework training, and training with manually annotated ground-truth bounding box (GT) labels.

\subsubsection{Challenges with SAM Pseudo-Labeling}

SAM has demonstrated impressive zero-shot segmentation capabilities across diverse 
 general-purpose datasets \cite{kirillov2023segment, yan2024segmentanythingmodelsachievezeroshot}. However, recent research \cite{2024adaptingsegmentmodelusage, Ji_2023, tang2023samsegmentanythingsam} has shown that SAM struggles in novel and specialized domains, such as aerial surveillance. Our experiments confirm this limitation: SAM often fails to produce meaningful segmentations for military vehicles when applied to UAV footage. The model frequently over-segments background textures while under-segmenting small, low-resolution objects of interest.

\begin{figure*}[h]
    \centering
    \begin{tabular}{ccc}
        \includegraphics[width=0.32\linewidth]{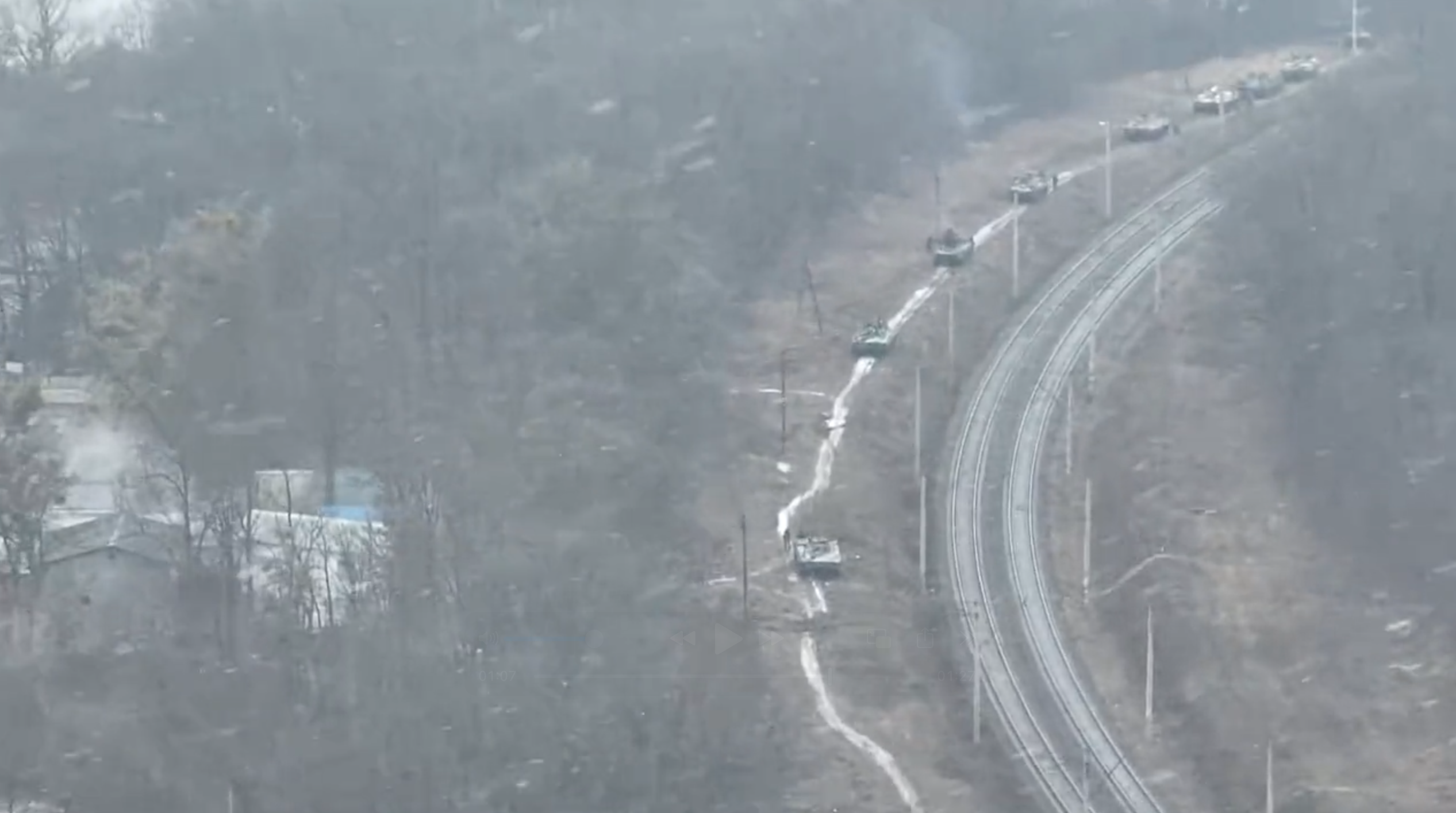} & 
        \includegraphics[width=0.32\linewidth]{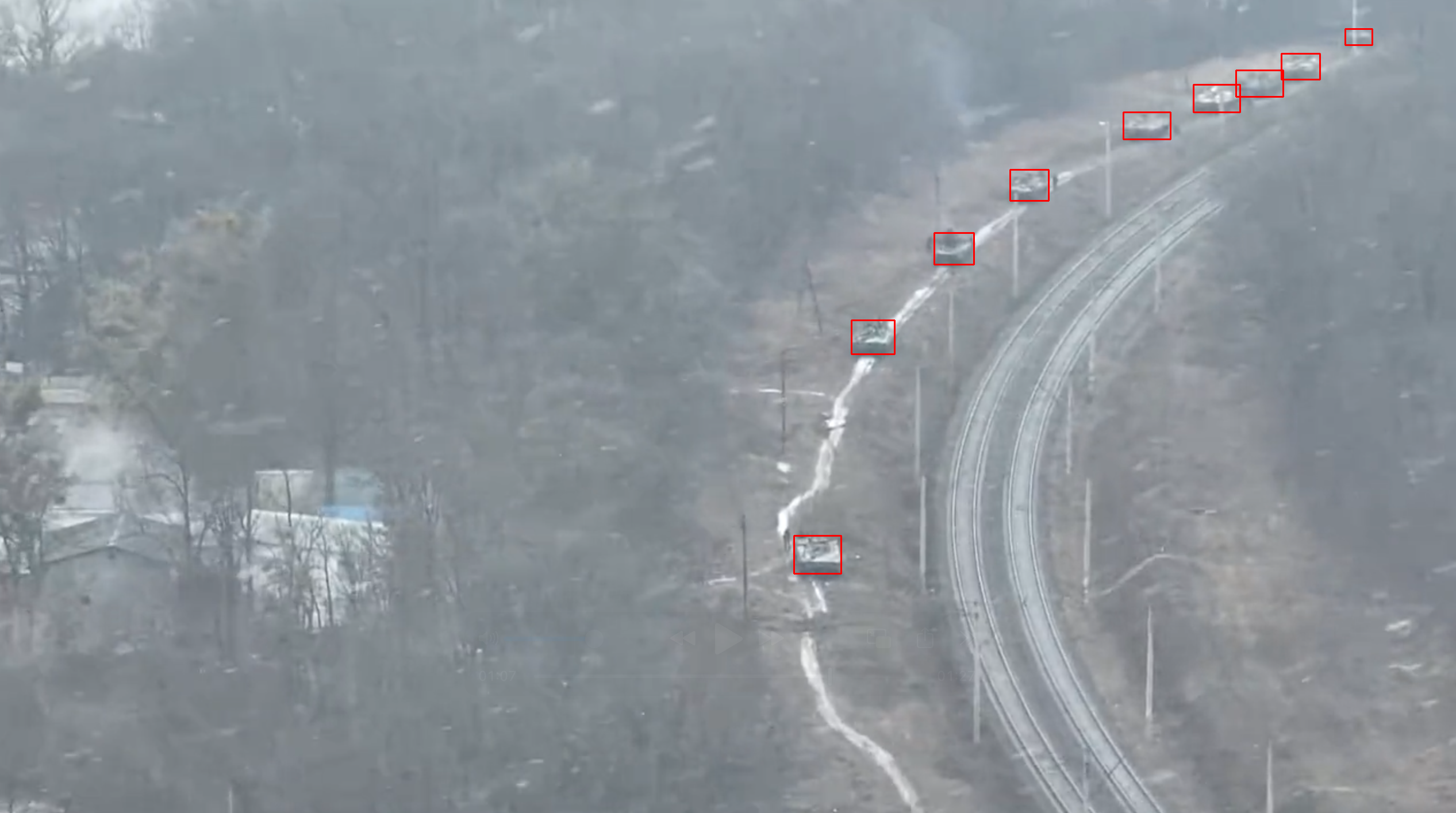} & 
        \includegraphics[width=0.32\linewidth]{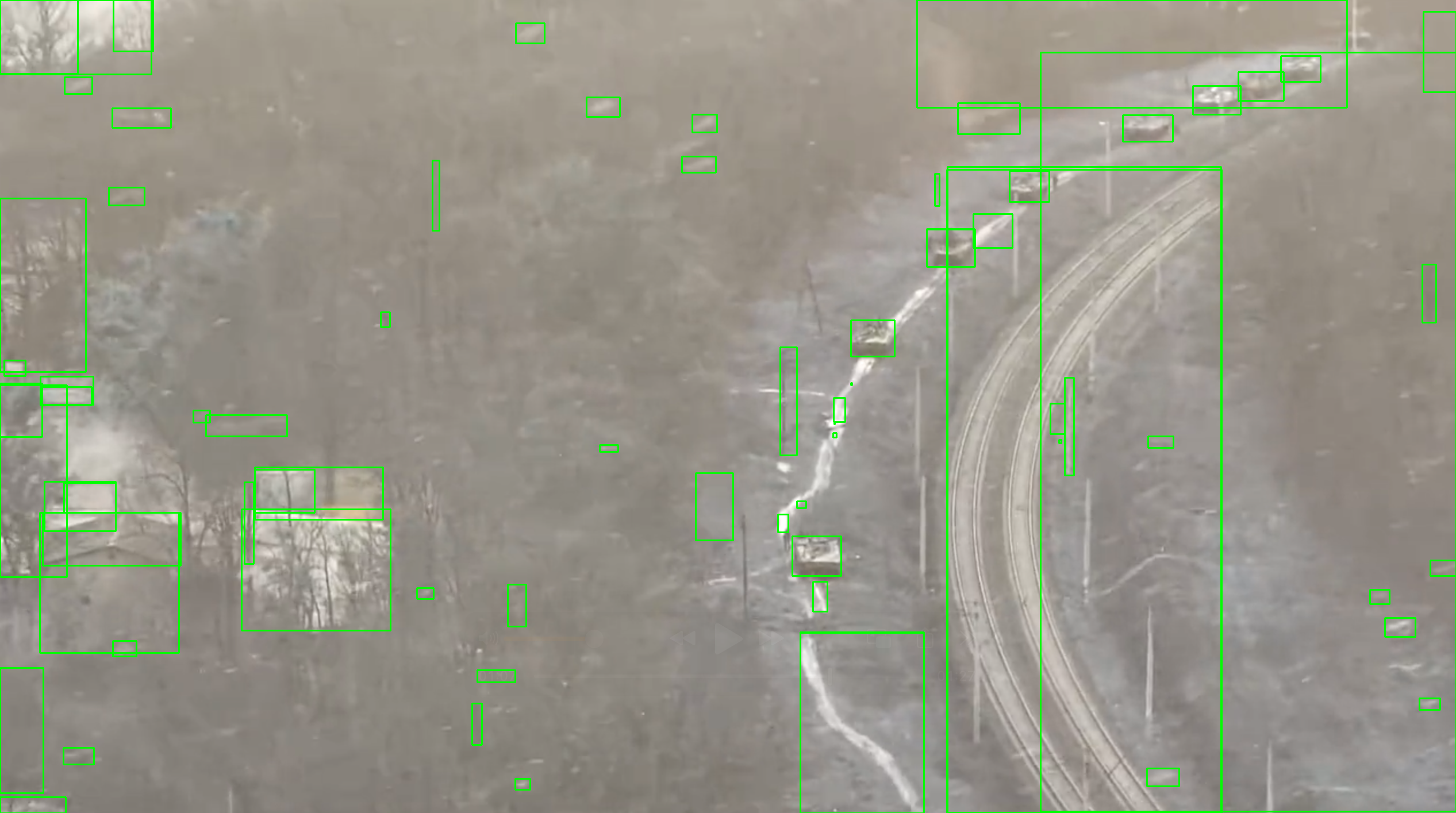} \\

        \includegraphics[width=0.32\linewidth]{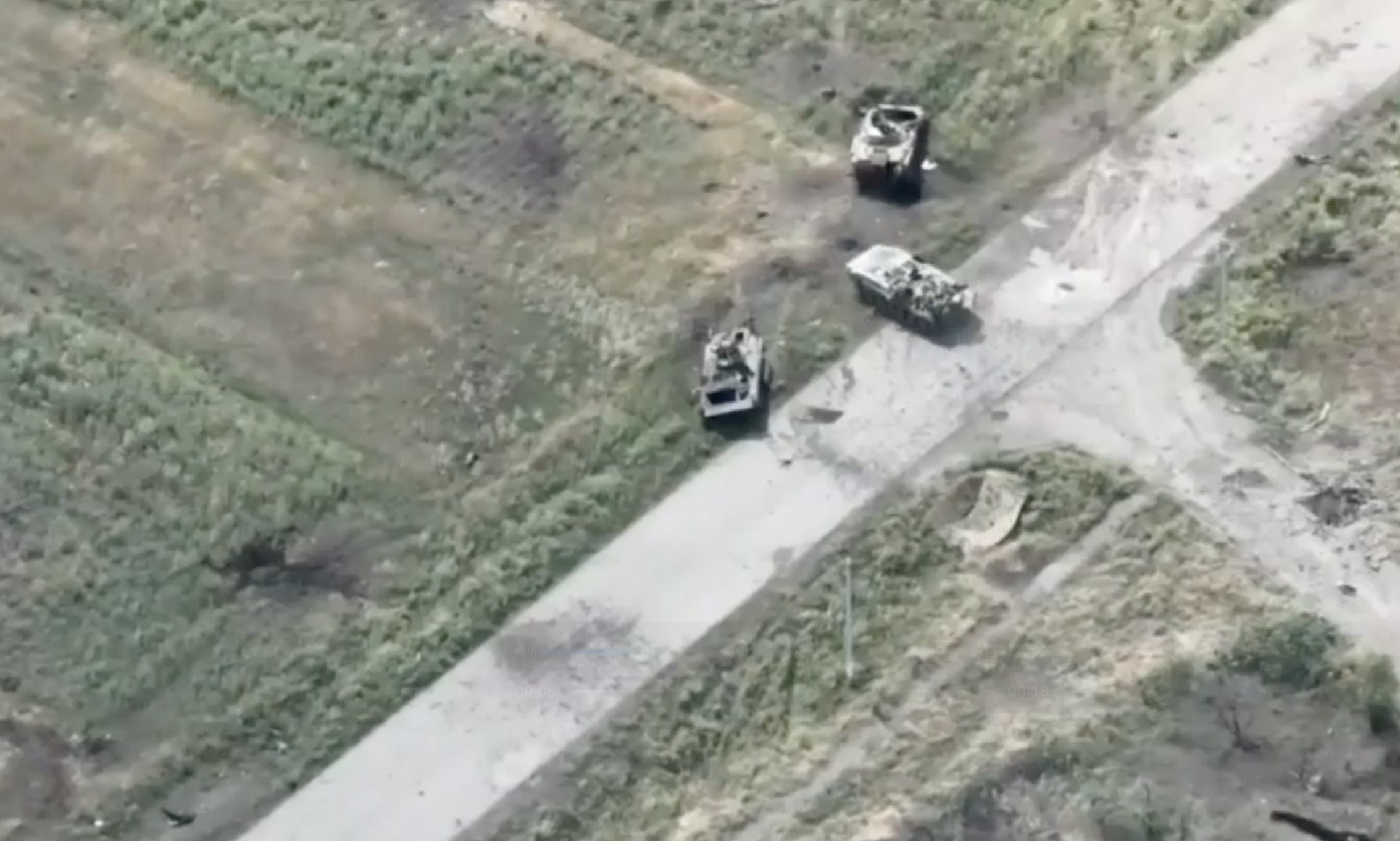} & 
        \includegraphics[width=0.32\linewidth]{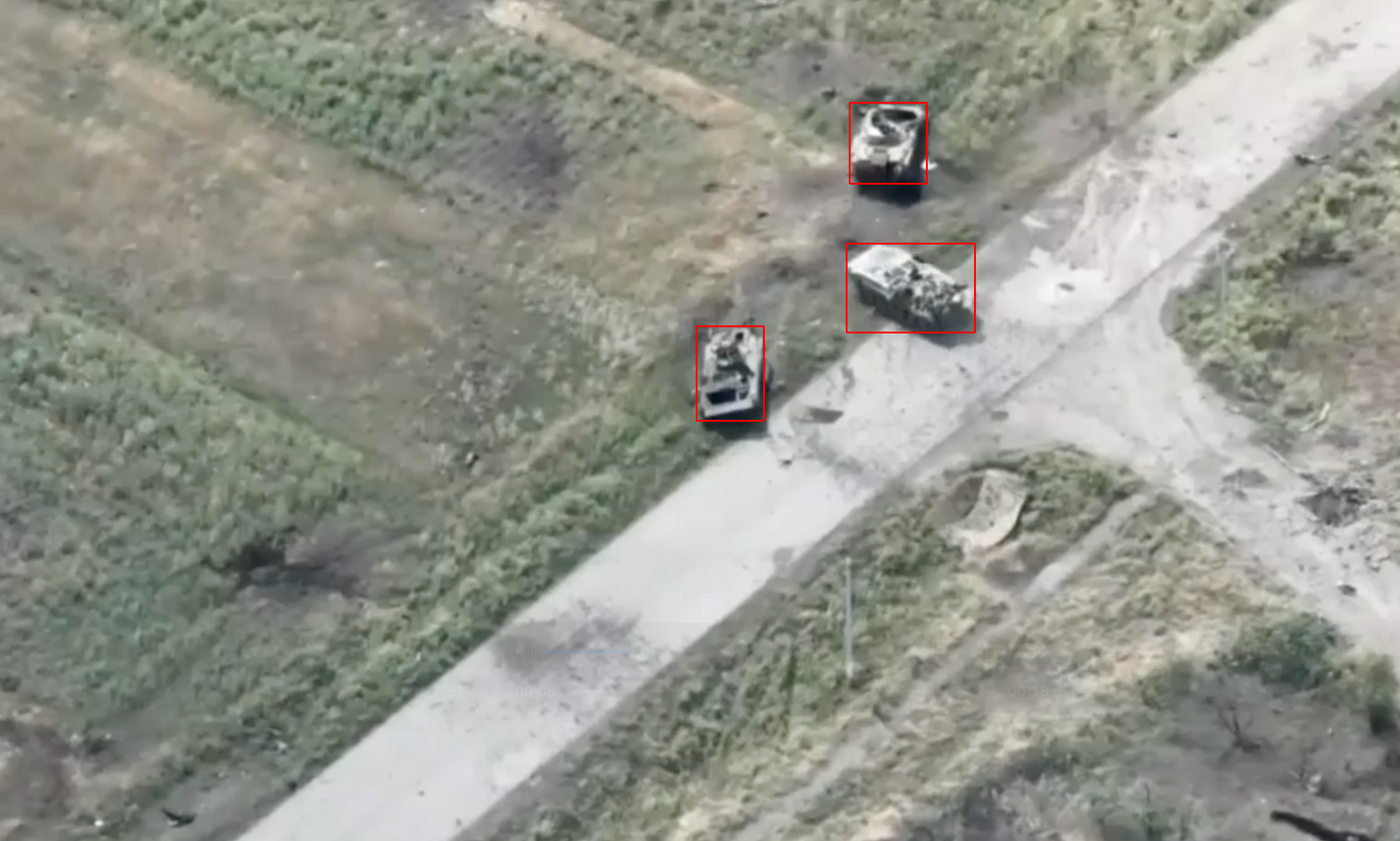} & 
        \includegraphics[width=0.32\linewidth]{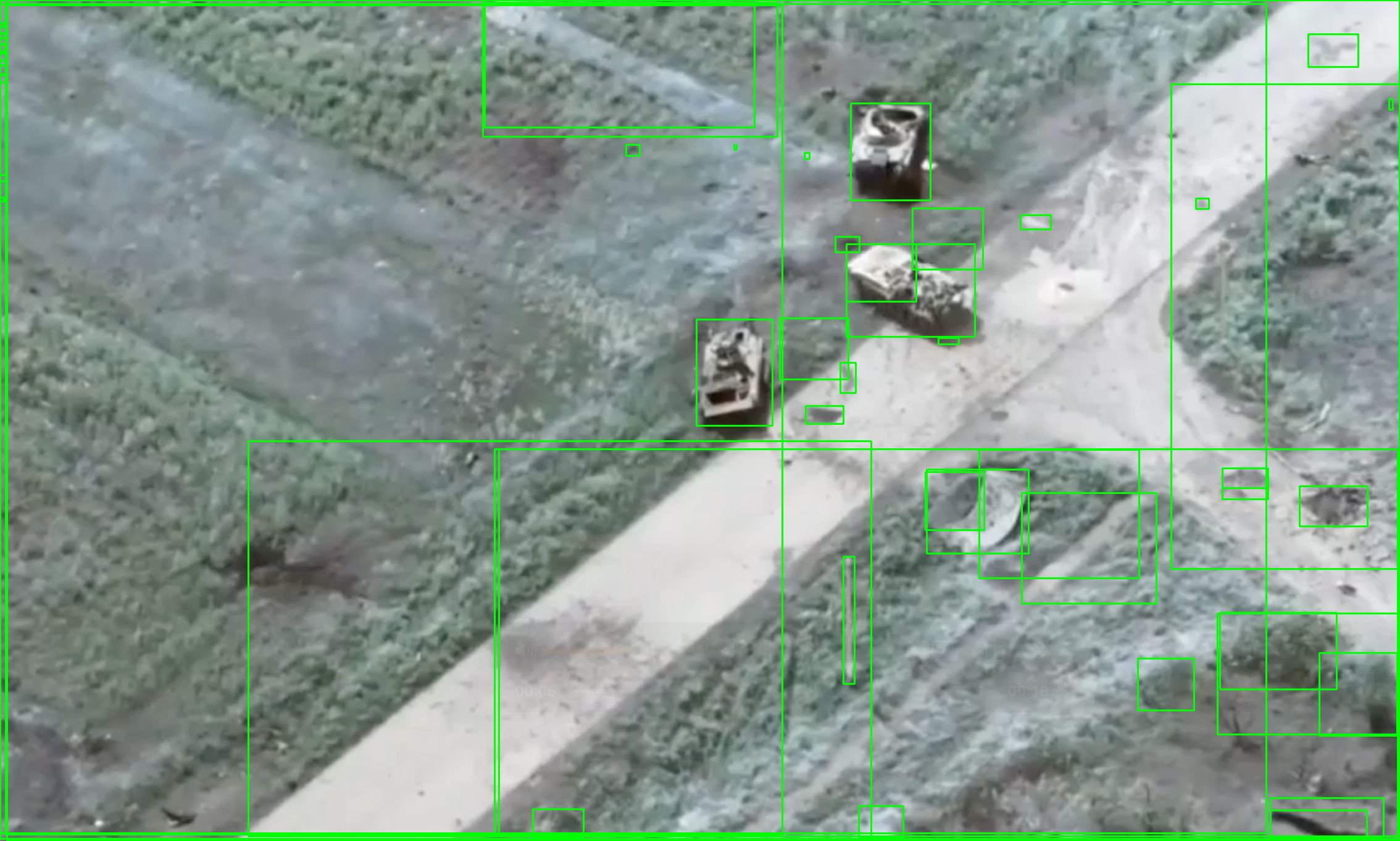} \\
        
        Raw Image & Ground Truth & SAM Annotation
    \end{tabular}
    \caption{Comparison of raw images (left), ground truth (middle), and SAM annotations (right). SAM struggles with precise segmentation, often mislabeling small military vehicles and over-segmenting environment textures.}
    \label{fig:sam_vs_gt}
\end{figure*}

Figure~\ref{fig:sam_vs_gt} illustrates these challenges by comparing SAM-generated pseudo-labels with manually annotated ground-truth bounding boxes. While SAM sometimes outlines vehicles correctly, its predictions are inconsistent and fail to distinguish between different military assets, leading to unreliable training supervision.

\subsubsection{Training on Ground-Truth Annotations}

Due to the limitations of SAM pseudo-labels, we opted to train our model using manually labeled ground-truth bounding boxes. Our data set comprises expert-annotated aerial footage, ensuring high-quality object localization. Training with precise GT annotations improved performance in both multiple-object and single-object tracking.

Table~\ref{tab:annotation_comparison} presents a quantitative comparison of model performance when trained on SAM pseudo-labels versus ground-truth labels. The results indicate a substantial improvement in HOTA and IDF1 scores when GT annotations are used.

\begin{table}[h]
    \centering
    \caption{Tracking Performance with Different Annotation Strategies}
    \label{tab:annotation_comparison}
    \begin{tabular}{|c|c|c|c|c|c|c|}
        \hline
        \textbf{Annotation Strategy} &  \textbf{HOTA} & \textbf{MOTA} & \textbf{IDF1} & \textbf{AssA} & \textbf{IDSW} \\
        \hline
        SA-1B-500K & 48.0 &	51.35 &	56.07 &	51.14 &	490 \\
        SAM Pseudo-Labels & 47.23 &	51.23 &	55.45 &	49.49 &	572 \\
        Ground-Truth Labels & \textbf{48.29} & \textbf{51.38} & \textbf{56.93} &	\textbf{51.80} & \textbf{471}\\
        \hline
    \end{tabular}
\end{table}

Our findings highlight that GT annotations provide significantly better results for tracking military vehicles in UAV footage. Leveraging precise bounding boxes ensures superior re-identification and tracking consistency, reinforcing the importance of high-quality annotation in low-FPS UAV-based military assets tracking tasks.

\subsection{Adaptability to Resolution Changes in Military Asset Tracking}

Military vehicles in UAV footage often appear as small objects that occupy only a few pixels in a frame. Their appearance changes rapidly due to the low frame rate of the footage, making traditional object-tracking approaches, which rely on temporal smoothness, ineffective. Additionally, military vehicles of the same type frequently exhibit nearly identical visual features, making it difficult to distinguish between them based solely on appearance. However, the broader scene context—such as terrain, nearby structures, and relative positioning—can provide crucial cues for re-identification.

To address these challenges, we train our MASA-based re-identification model on frames with a maximum size of 1280 resolution but perform inference on multiple downscaled resolutions: 640, 320, and 160. This scaling strategy simulates real-world constraints where computational efficiency may require lower-resolution processing. Despite the drastic reduction in object size during inference, the model maintains high re-identification accuracy. The learned feature embeddings leverage not only the object's appearance but also its surrounding context, which remains informative even at lower resolutions.

Notably, the reliability of our approach comes from the way MASA learns instance associations by pooling the features from the frame-level context by ROI-Align operation \cite{he2018maskrcnn}. Instead of relying purely on fine-grained details of individual objects, the model extracts global and local contextual features that remain stable across different scales. As a result, even when objects become tiny (with an area of less than 32 pixels) after downscaling, their relative position within the environment enables tracking to function without a significant drop in performance. This shows that incorporating spatial context is crucial for military asset tracking in low-FPS UAV footage, mainly when dealing with visually similar targets.

Our experimental results confirm that the performance drop due to resolution reduction is minimal. Re-identifying objects effectively at lower resolutions enables real-time tracking while reducing computational overhead, making our approach practical for large-scale UAV surveillance applications.

\begin{table}[h]
    \centering
    \caption{Tracking Performance at Different Resolutions}
    \label{tab:resolution_performance}
    \begin{tabular}{|c|c|c|c|c|c|c|}
        \hline
        \textbf{Resolution} & \textbf{HOTA} & \textbf{MOTA} & \textbf{IDF1} & \textbf{AssA} & \textbf{IDSW}  \\
        \hline
        1280, 736 & \textbf{48.29} & 51.38 & \textbf{56.93} &	\textbf{51.80} & \textbf{471}  \\
        640, 368  & 48.19 & \textbf{51.40} & 56.71	 & 51.53 & 472  \\
        320, 184  & 48.02 & 51.23 & 56.37 & 51.27 & 510  \\
        160, 92  & 46.57 & 51.30 & 54.14 & 48.06 & 579  \\
        \hline
    \end{tabular}
\end{table}

Table~\ref{tab:resolution_performance} presents the tracking performance at different resolutions. Even as the input resolution decreases, the HOTA and IDF1 scores remain relatively stable, confirming that contextual information plays a key role in object association and tracking. The minimal drop in performance at lower resolutions highlights the effectiveness of leveraging global and local context for robust re-identification in UAV-based military asset tracking.

\subsection{Latent Representation Dimensionality}

The dimensionality of the object's latent representation plays a crucial role in balancing model performance and computational efficiency. The original model used a latent representation size of 256, providing a strong baseline for tracking accuracy and association quality. However, inspired by recent work on contrastive learning of visual representation \cite{contrastive_2020, chen2020simpleframeworkcontrastivelearning}, we observed that decreasing the latent dimensionality to 64 and even 32 resulted in no significant drop in key evaluation metrics, including HOTA and IDF1.

Reducing the embedding size led to notable improvements in inference speed and memory footprint, making the model more practical for real-time deployment, particularly in low-resource edge environments such as UAV-based military tracking systems.

To quantify the impact of embedding size reduction, we compare the performance of our model at different embedding dimensions in Table~\ref{tab:embedding_comparison}.

\begin{table}[h]
    \centering
    \caption{Comparison of model performance with different embedding sizes}
    \label{tab:embedding_comparison}
    \begin{tabular}{|c|c|c|c|c|c|c|c}
        \hline
        \textbf{Res.} & \textbf{Emb. Size}  & \textbf{HOTA} & \textbf{MOTA} & \textbf{IDF1} & \textbf{AssA} & \textbf{IDSW}  \\
        \hline
        1280 & 256 & 48.29 & 51.38 & 56.93 & 51.80 & 471 \\
        1280 & 64  & 48.00 & 51.47 & 56.54 & 51.07	& 448 \\
        160  & 32  & 46.62 & 51.03 & 54.18 & 48.3 & 645 \\
        \hline
    \end{tabular}
\end{table}

Further optimization efforts focused on developing a faster-performing model for large-scale tracking scenarios. This optimized version operates at an image resolution of 160 while maintaining an embedding size of 32, striking an optimal balance between computational efficiency and tracking quality. Despite the drastic reduction in input resolution and latent feature dimensionality, the model continues to achieve formidable tracking performance, demonstrating the effectiveness of our approach in lightweight object association and re-identification tasks. The results are shown in Table ~\ref{tab:embedding_comparison}.

Our findings highlight that tracking performance remains highly resilient to reductions in latent dimensionality, suggesting that even lower-dimensional embeddings retain sufficient discriminative power for instance association. This insight opens avenues for further exploration in lightweight tracking models tailored for low-power devices and real-time intelligence applications.

\section{Conclusions}

In the current work, we explored how instance association learning from single-frame annotations can be used to overcome typical problems posed by the task of real-time tracking of military
vehicles in low-fps UAV footage. We show how a self-supervised framework can be used for the task, even when ground truth data doesn't contain object tracking information. Our experiments show that single-frame ground truth bounding boxes are superior to pseudo-labels generated with the SAM model as a source of training data for tracking models in the context of military assets tracking from UAVs. 

Our results also suggest that the reduction of an input image resolution and the dimensionality of the object’s latent representation can be performed without a significant drop in the object re-identification accuracy. These findings open a possible avenue for optimization of the tracking model's memory footprint and the demand for computational resources during inference.

Despite the clear advantage of utilizing only a single-frame ground truth data (no object tracking information in GT) for training, which dramatically reduces the effort required for the training data annotation, the approach presented here has some limitations. One of them is an absence of long-term memory. If an object leaves the camera's field of view for more than a couple of seconds, the model struggles with building a robust temporal association for the object. Another limitation is that it can be hard for the model to efficiently associate the same object with itself if viewed from multiple different points of view simultaneously, e.g. from two different UAVs.

We expect further improvements in multiple object tracking will provide efficient solutions to the above mentioned limitations. A publicly available, curated dataset for real-time MOT of military
vehicles in low-FPS UAV footage plays a key role in tracking this progress. Here, we establish such a dataset, which is available upon request through   \href{https://docs.google.com/forms/d/e/1FAIpQLSdKceLoF_L3242FDJLLFdIiVDepQ0imt54SwbxnFgt1sSpjug/viewform?usp=sharing}{\underline{this form}}.

\section*{Acknowledgment}

We extend our deepest gratitude to the Armed Forces of Ukraine for their unwavering courage and resilience in defending our country. Their dedication and sacrifice inspire us all. We also wish to thank the CIDTD and UCU teams for their steadfast support, guidance, and collaborative spirit, which have been instrumental in developing this work. 

ELEKS supported this project through a grant dedicated to the memory of Oleksiy Skrypnyk. The research was also partially supported by the National Research Foundation of Ukraine (grant no. 2023.04/0158). 
\bibliographystyle{plain}   % or another style like ieee, apalike, etc.
\bibliography{references}   % <-- name of your .bib file without the .bib extension

\vspace{12pt}

\end{document}